\documentclass[journal]{IEEEtran}
\usepackage[cmex10]{amsmath}
\usepackage{amssymb}
\usepackage{amsfonts}
\usepackage{amsthm}
\usepackage{algorithm}
\usepackage{algpseudocode}
\usepackage{graphicx}
\usepackage{epsfig}
\usepackage{gensymb}
\usepackage{hyperref}
\hypersetup{
	colorlinks=true,
	linkcolor=blue,
	filecolor=magenta,      
	urlcolor=cyan,
}
\usepackage{cite}
\usepackage{tensor}
\usepackage{colortbl}
\usepackage[caption=false,font=footnotesize]{subfig}
\usepackage{color}
\input{abkuerzung_IEEE.sty}
\graphicspath{{img/}}
\usepackage{bm}
\usepackage{url}
\usepackage{dirtytalk}
\usepackage{todonotes}
\usepackage{fancyhdr}

\DeclareGraphicsExtensions{.eps}
\theoremstyle{definition}

\DeclareMathAlphabet\mathbfcal{OMS}{cmsy}{b}{n}
\usepackage{epigraph}

\renewcommand{\baselinestretch}{0.98}

\begin{document}
	
	\title{Challenges and Outlook in Robotic Manipulation\\of Deformable Objects}
	
	\author{Jihong~Zhu,~Andrea~Cherubini,~Claire~Dune,~David~Navarro-Alarcon,~Farshid~Alambeigi,~Dmitry~Berenson,~Fanny~Ficuciello,~Kensuke~Harada,~Jens~Kober,~Xiang~Li,~Jia~Pan,~Wenzhen~Yuan and Michael~Gienger%
		\thanks{J. Zhu is with Cognitive Robotics, TU Delft and Honda Research Institute, Europe. {\tt\small j.zhu-3@tudelft.nl}}%
		\thanks{A. Cherubini is with LIRMM - Universit\'{e} de Montpellier CNRS, 161 Rue Ada, 34090 Montpellier, France. {\tt\small andrea.cherubini@lirmm.fr}}
		\thanks{C. Dune is with COSMER laboratory EA 7398, Universi\'{e} de Toulon,
			83130 La Garde, France. {\tt\small claire.dune@univ-tln.fr}}
		\thanks{D. Navarro-Alarcon is with The Hong Kong Polytechnic University, Department of Mechanical Engineering, Kowloon, Hong Kong. {\tt\small dna@ieee.org}}
		\thanks{F. Alambeigi is with the  University of Texas at Austin, Austin, USA. {\tt\small farshid.alambeigi@austin.utexas.edu}}
		\thanks{D. Berenson is with the University of Michigan, Ann Arbor, MI, USA. {\tt\small berenson@eecs.umich.edu}}
		\thanks{F. Ficuciello is with Università degli Studi di Napoli Federico II, 80125
			Napoli, Italy {\tt\small fanny.ficuciello@unina.it}}
		\thanks{K. Harada is with the Osaka University, Japan, and the National Institute of AIST, Japan {\tt\small harada@sys.es.osaka-u.ac.jp}}
		\thanks{J. Kober is with Cognitive Robotics, TU Delft, the Netherlands {\tt\small J.Kober@tudelft.nl}}
		\thanks{X. Li is with Department of Automation, Tsinghua University, Beijing, China. {\tt\small xiangli@tsinghua.edu.cn}}
		\thanks{J. Pan is with with the Department of Computer Science, The University of
			Hong Kong, Pok Fu Lam, Hong Kong. {\tt\small jpan@cs.hku.hk}}
		\thanks{W. Yuan is with Robotics Institute, Carnegie Mellon University, Pittsburgh, PA 15213 USA {\tt\small wenzheny@andrew.cmu.edu}}
		\thanks{M. Gienger is with Honda Research Institute Europe, Offenbach, Germany {\tt\small Michael.Gienger@honda-ri.de}}
		\thanks{This work has been submitted to the IEEE for possible publication. Copyright may be transferred without notice, after which this version may no longer be accessible.}
	}
	
	\bstctlcite{IEEEexample:BSTcontrol}
	
	\markboth{Zhu \MakeLowercase{\textit{et al.}}: ArXiv Preprint, conditionally accepted for IEEE RA-M}%
	{Zhu \MakeLowercase{\textit{et al.}}: Challenges and Outlooks in Deformable Object Manipulation}
	
	\maketitle
	\begin{abstract}
		Deformable object manipulation (DOM) is an emerging research problem in robotics. The ability to manipulate deformable objects endows robots with higher autonomy and promises new applications in the industrial, services, and healthcare sectors. However, compared to rigid object manipulation, the manipulation of deformable objects is considerably more complex, and is still an open research problem. Addressing DOM challenges demand breakthroughs in almost all aspects of robotics, namely hardware design, sensing, (deformation) modeling, planning, and control. In this article, we review recent advances and highlight the main challenges when considering deformation in each sub-field. A particular focus of our paper lies in the discussions of these challenges and proposing future directions of research.
		
	\end{abstract}
	
	
	
	%
	\IEEEpeerreviewmaketitle
	\section{Introduction}
	\IEEEPARstart{U}{ntil} now, object rigidity is one of the common assumptions in robotic grasping and manipulation. Strictly speaking, all objects deform upon force interaction. Rigidity is a valid assumption when object deformation can be neglected in the task. Nevertheless, many objects that need to be manipulated by robots present non-negligible deformation: from micro surgical operation to challenging industrial assemblies.
	
	Robots need to be capable of manipulating deformable objects to operate in human environments. This capability would benefit many application fields, while posing fundamental research challenges. In this article, we consider a generalized concept of manipulation where grasping is also part of the task. We will refer to the problem as deformable object manipulation (DOM).
	
\begin{figure}[t]
		\centering
		\includegraphics[width=0.48\textwidth]{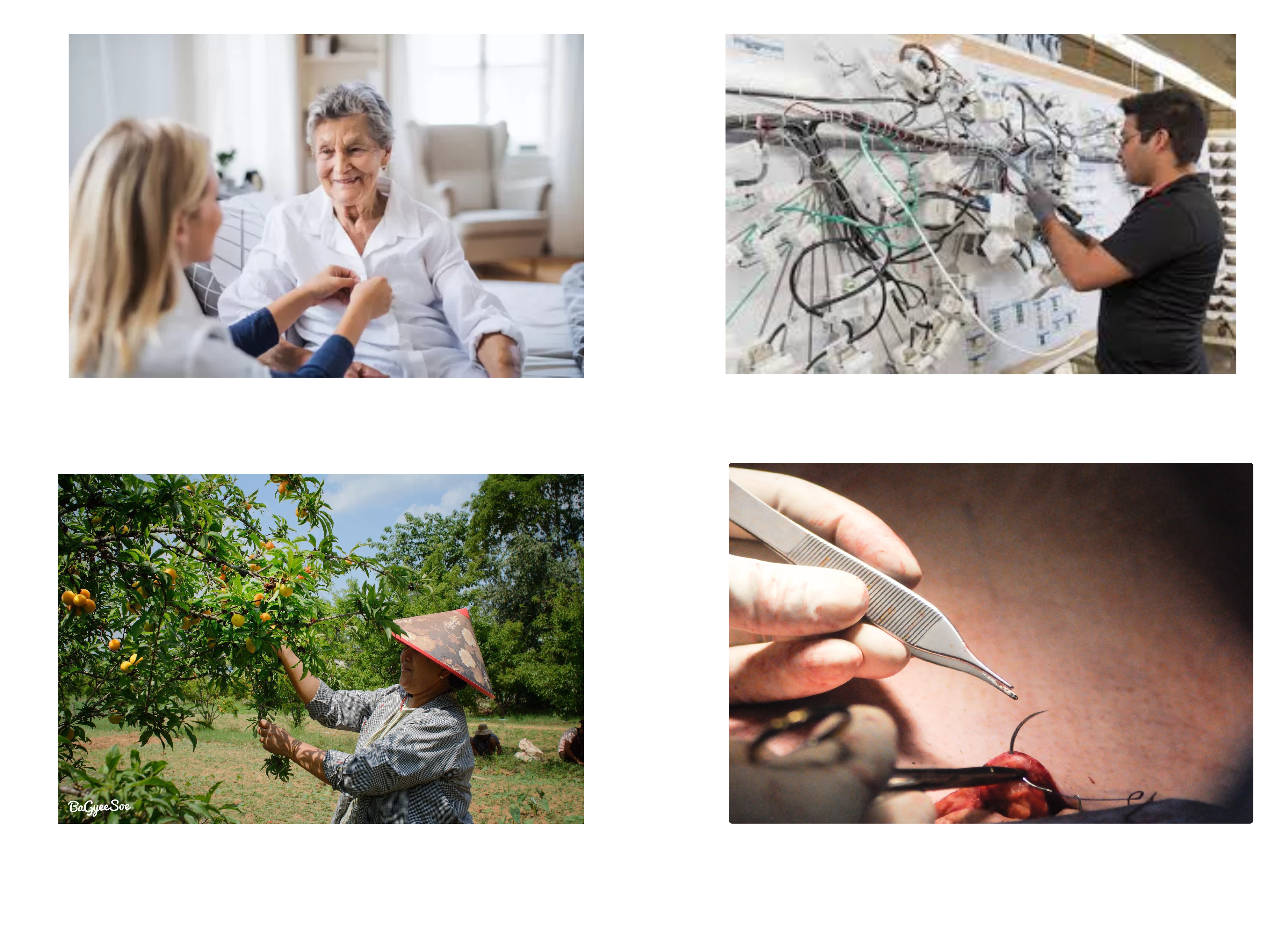}
		\caption{Applications involving manipulation of deformable objects. Clockwise from top left: dressing assistance~\cite{assist_dressing}, cable harnessing~\cite{cable}, fruit harvesting~\cite{fruit_harvesting}, suturing~\cite{suture}}
		\label{fig:real-world-applications}
	\end{figure}

	The tasks involved in DOM cover a broad spectrum (see Fig. \ref{fig:real-world-applications}). These include: dressing assistance in elderly care, cable harnessing in industrial automation, harvesting and processing fruit and vegetables in agriculture, surgical operations in medical services, to name a few.
	
	On the technical side, addressing deformation introduces the following technical challenges:
	\begin{itemize}
		\item the complication of sensing deformation,
		\item the high number of degrees of freedom of soft bodies,
		\item the complexity of non-linearity in modeling deformation.
	\end{itemize}
	
	We believe that overcoming these challenges is not only beneficial to DOM, but can further push towards developing autonomous robots which can operate in unstructured environments.
	\begin{figure*}[t!]
		\centering
		\includegraphics[width=0.6\textwidth]{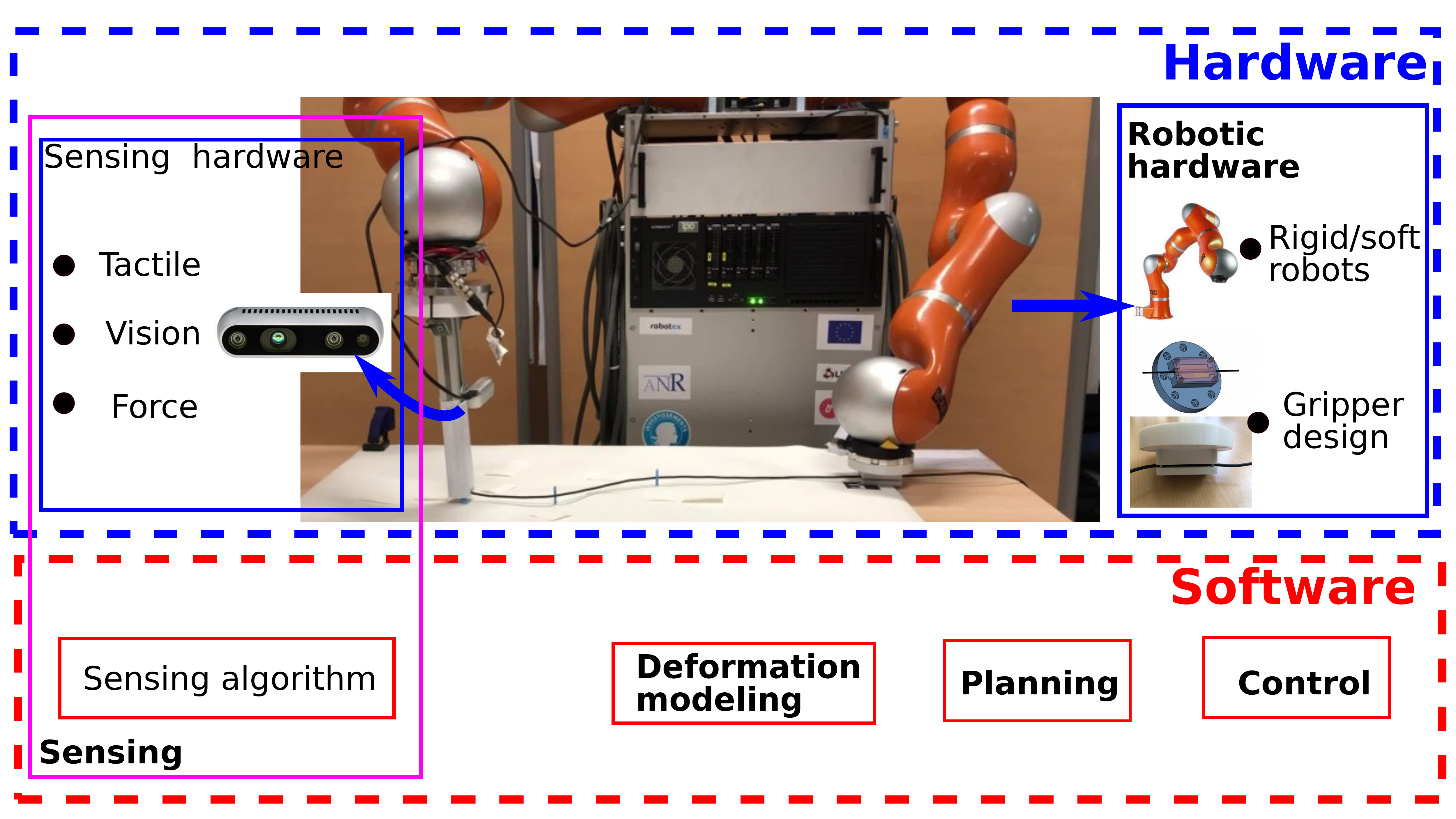}
		\caption{A typical robotic framework for handling deformable objects. In this particular example, the framework addresses wire harness~\cite{zhu2019robotic}.}
		\label{fig:system-level-overview}
	\end{figure*}
	In recent years, there have been a few surveys on robotic manipulation of deformable objects.
	Some surveys focus on specific areas of DOM.
	The survey from Jimenez \cite{jimenez2012survey} focuses on model-based manipulation planning. More recently, Herguedas et al. \cite{herguedas2019survey} review works using multi-robot systems for DOM while the work of \cite{nadon2018multi} considers multi-modal sensing. The authors of \cite{arriola2020modeling} present the state-of-the-art on deformable object modeling for manipulation.
	There are also two comprehensive surveys in the area.
	The survey in \cite{sanchez2018robotic} reviews and classifies the state-of-the-art according to the object's physical properties. Lately, \cite{yin2021modeling} reports most recent advances in modeling, learning, perception, and control in DOM. 
	
	In contrast with the mentioned surveys, which either focus on reporting the progress of the field or a specific area, this article aims at identifying \emph{scientific challenges} introduced by object deformations and at projecting crucial future research directions. As DOM is an emerging field of research where there is still much to be done, in this paper, previous works and open problems are given equal weights. In addition, we dedicate one section to discussing practical challenges in various applications of DOM.  
	We believe the paper is a first of its kind, in the field of DOM.
	
\begin{figure}[h!]
		\centering
		\subfloat[Highest qualifications of the respondents.]{\includegraphics[width=0.3\textwidth]{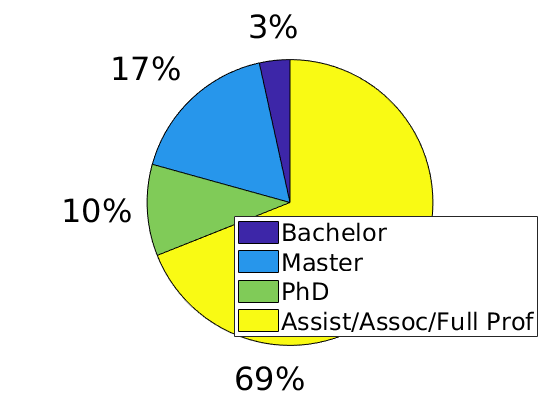}\label{fig:qualifications}}
		\hspace{2mm}
		\subfloat[Means and variances of Importance and research maturity ratings of each key component.]{\includegraphics[width=0.5\textwidth]{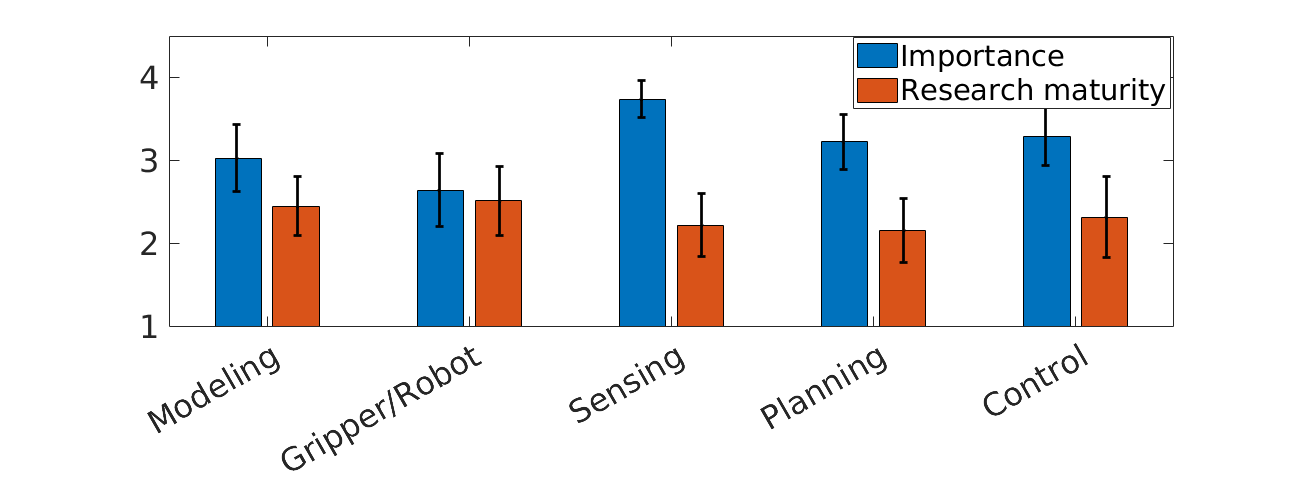}}
		\caption{Summary of the outcomes of the survey on DOM. We received in total $31$ answers. The respondents cover different level of qualifications ranging from master students to full professors.}
		\label{fig:maturity_importance}
	\end{figure}

	A robotic framework designed to handle deformable objects usually consists of five key components: \emph{gripper and robot design}, \emph{sensing}, \emph{modeling}, \emph{planning} and \emph{control} (See Fig. \ref{fig:system-level-overview}). To position the current research and identify future trends, we conducted a survey on the future perspective of deformable object manipulation\footnote{Link to the survey: \url{https://forms.gle/XCv2CV79yvRP5Gsd7}}. We shared the survey with people working in related field, at various career stages. They were asked to rate the importance and research maturity of each of the five identified key components, from $1$ to $4$, with $1$ being not important/low maturity and $4$ being very important/high maturity. We received 31 answers that are summarized on Fig. \ref{fig:maturity_importance}.
	
	
	We consider the promising direction of research as the ones that have the highest significance and the lowest research maturity. Based on the survey, sensing is the most promising one among all subareas. This is probably due to current booming trend in Deep Learning has offered many new methods for processing the sensory data. In addition, sensing is the prerequisite for subsequent steps such as modeling, planning and control.  
	
	Accordingly, the following sections of this paper each present one of these five research directions. In each section, we review recent works in the field and then comment the outlook and challenges ahead. Then, Sect. \ref{sec:application} tries to provide a link from research to practical applications in the context of DOM.  Finally, we summarize key messages in Sect. \ref{sec:summary}.
	
	
	\section{Gripper and robot design}\label{sec:hardware} 
	\subsection{Current capability}\label{sec:hardware_current}
	
\begin{figure}[!thpb]
		\centering
		\includegraphics[width=0.48\textwidth]{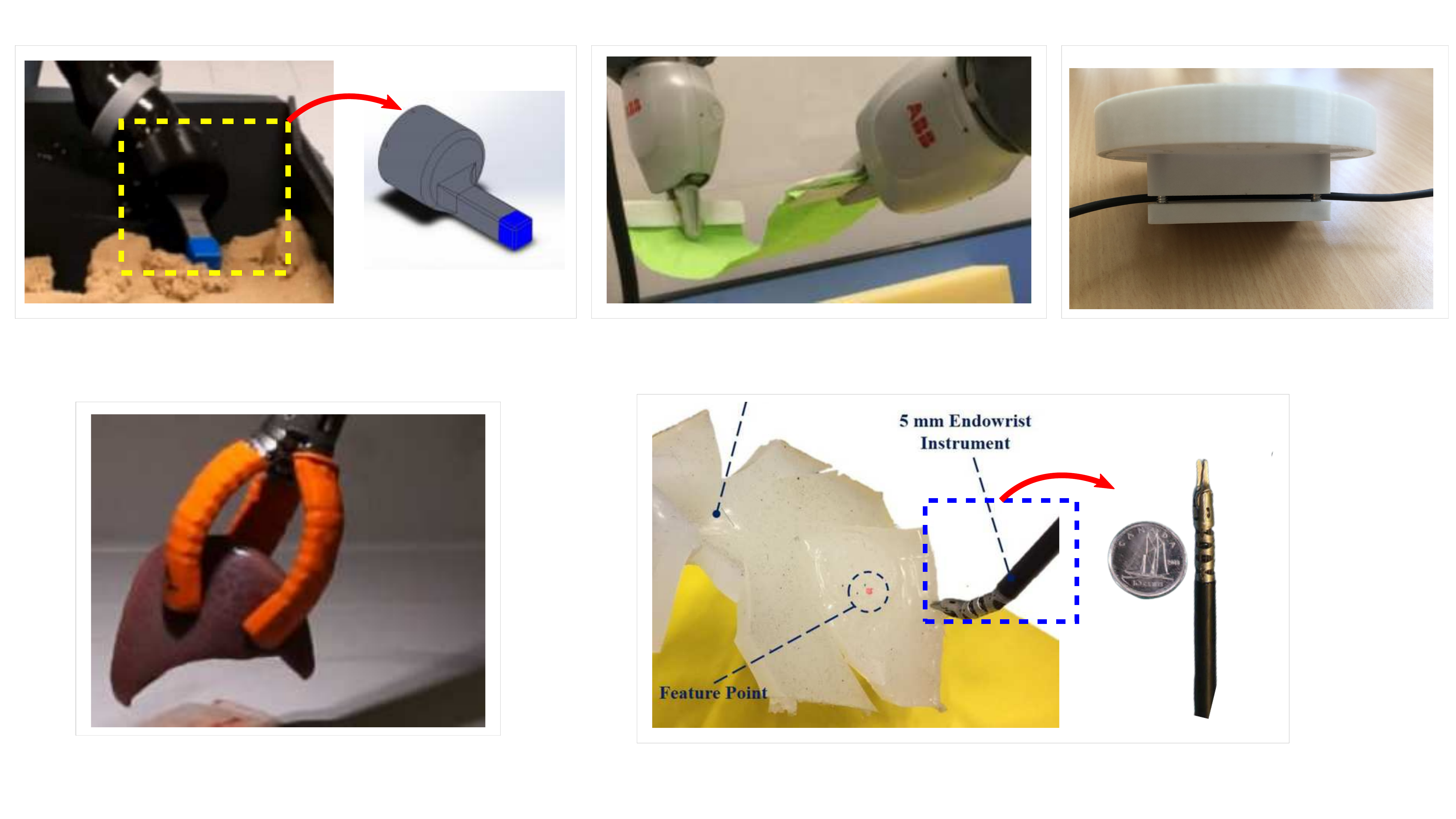}
		\caption{Various robot grippers for DOM. Clockwise from top left: a tool for pushing and tapping on plastic materials \cite{cherubini2020model}, flat clips for holding a towel \cite{hu20193}, a gripper allowing a cable to slide \cite{zhu2019robotic}, a soft continuum manipulator interacting with a deformable material  \cite{alambeigi2018autonomous} and a soft hand for manipulating organs \cite{liu2020musha}.}
		\label{fig:hardware}
	\end{figure}

	Does manipulation of deformable objects demand specific grippers as compared to manipulation of rigid objects? Generally, yes (see Fig. \ref{fig:hardware}). Unlike rigid objects (which are mostly handled by standard grippers), deformable objects are handled with custom (and often designed ad-hoc) grippers, e.g. a 3D printed gripper that enables cable sliding \cite{zhu2019robotic}, a flat clip for holding towels \cite{hu20193}, a cylindrical tool for pushing and tapping plastic materials \cite{cherubini2020model}, a soft hand for manipulating organs \cite{liu2020musha}.
	Such diversity in grippers is a result of the large variety of deformable objects, which require different actions during manipulation. To avoid designing task-specific grippers for DOM, human-like dexterity and compliance is desired. Recent works in this directions consider compliant design \cite{della2015dexterity, abondance2020dexterous} and show good potential for DOM tasks. 
	
	As for the robot itself, it is rigid in most works. In some cases, as in the surgical application showcased in~\cite{alambeigi2018autonomous} (Fig. \ref{fig:hardware}, bottom right), both robot and object are deformable to ensure safety of manipulation. 
	
	\subsection{Challenges and outlook}\label{sec:hardware_outlook}
	Improving dexterity is core to robot manipulation. The improvement can come from different research domains, such as accurate in-hand sensing or robust control, two aspects which we will detail in Sections \ref{sec:sensing} and \ref{sec:control}, respectively. In this section, our focus is on gripper/robot hardware aspects.
	
	One way of achieving such dexterity is to reproduce by design the most dexterous gripper -- the human hand. An open question is whether anthropomorphic design is in itself the optimal solution in all cases, especially in the context of DOM.  
	
	While having one dexterous gripper which can handle a variety of DOM tasks is appealing, it should be noted that additional constraints need to be considered in the design process, for hygiene/safety in tasks such as food handling or surgery. For instance, for surgical applications we are limited by biocompatibility of the materials and actuators and by the reduced available space in minimally invasive surgery. In theses cases, designing task-specific grippers is more appropriate. Non-anthropomorphic soft grippers are another emerging area of research \cite{Hao2021}. These grippers are promising, to overcome the challenges associated with traditional fingered grippers in grasping rigid objects; yet, to date, their application to DOM receives little attention.
	
	Otherwise, one may use a standard gripper, and provide the robot with suitable tools to be grasped and used according to the type of task at stake. This demands breakthroughs on the algorithmic side, to make the robot capable of reasoning on the proper tools for different tasks. Training the robot to have task-specific reasoning will enhance autonomy and make robots realize more complex tasks.
	
	\begin{figure}[b]
		\centering
		\subfloat[Picking raspberries~ \cite{strawberry-picking}.]{\includegraphics[width=0.215\textwidth]{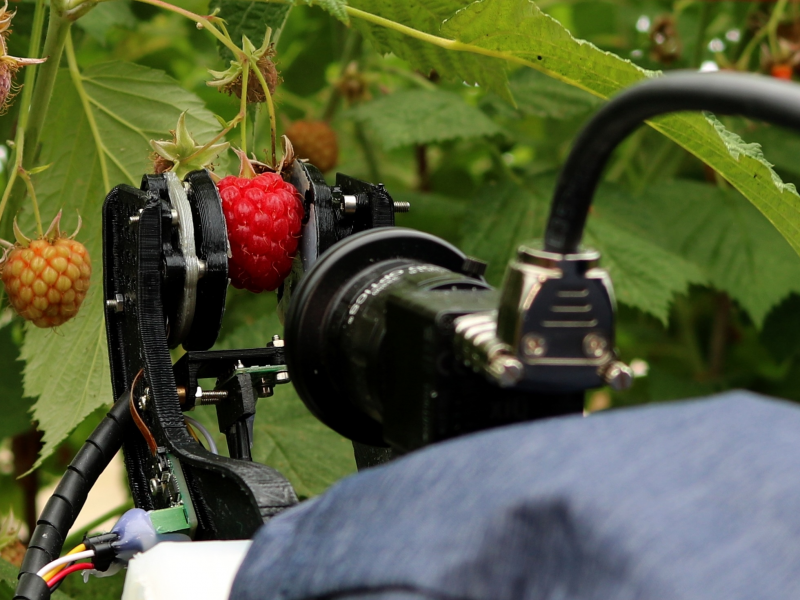}\label{fig:strawberry}} \hspace{2mm}
		\subfloat[A custom 3D printed soft robotic gripper, grasping mushroom coral~\cite{Vogt2018-po}]{\includegraphics[width=0.215\textwidth]{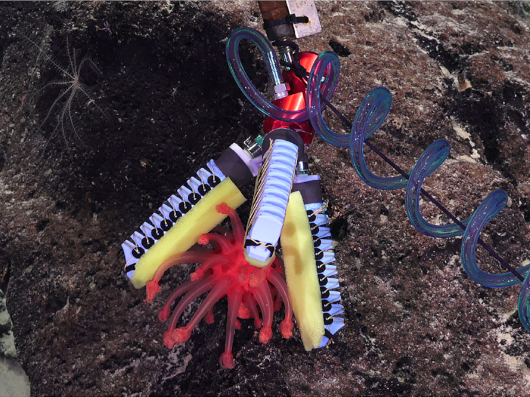}\label{fig:coral_reef}}
		\caption{Two examples of interaction with fragile objects, which could benefit from the use of soft robots.}
		\label{fig:soft-robots}
	\end{figure}
	
	Another area worth investigating is that of soft robots/grippers, since these have great potential for manipulating fragile materials, such as organs or food, or for collecting biological samples or fruits (see Fig. \ref{fig:soft-robots}). While traditional rigid robots need to exhibit compliant behavior when interacting with these objects, the inherent compliance of soft robots makes the task safe. This unconventional paradigm of using soft robots to manipulate soft objects will bring new challenges in modeling and control as both the robot and the object are under-actuated and difficult to model. One pioneering work in this direction is \cite{Ficuciello2018}, which adapts the finite element modeling (FEM) based inverse soft robot model with contact handling (proposed in \cite{Coevoet2017}) for deformable objects manipulation using soft robots.
	
	An interesting research question to consider is whether methods can be transferred from one field to the other. To be more specific, can methods for controlling/modeling soft robots be applied to manipulating deformable objects and vice versa? If so, as a community, it may be valuable to obtain a unified approach for working with both soft robots and deformable objects.
	
	
	\section{Sensing}\label{sec:sensing}
	
	\subsection{Current capability}\label{sec:sensing_current}
	
	In this section, we consider \textit{visual}, \textit{tactile} and \textit{force} sensing for DOM. Existing research relies on these three modes to estimate the state of deformable objects. In most cases, vision provides global information about shapes on a large scale, while force and tactile provide local information on both shape and contact. At the end of this section, we also discuss the research in contrast with this common practice, where global deformation properties are recovered using tactile sensing. It should also be noted that force information is particularly important in industrial settings, e.g., for assembly \cite{luo2018deep}, \cite{hayami2019multi}. 
	Vision is used in tasks such as rope manipulation \cite{nair2017combining,yan2020self} or cloth unfolding \cite{sun2015accurate,li2015regrasping}, where the object exhibits large global deformation. In these works, configurations of deformable objects were obtained from raw image readings. Although vision offers a global perspective of the object configuration, visual data can be noisy in unstructured environments, it is then important to manage occlusions~\cite{hu20193,cheng2019occlusion}.  
	Most of above-mentioned works are based on 2D vision, 3D perception of deformable objects is more challenging. Existing works employ FEM~\cite{petit2017tracking} or a combination of Growing Neural Gas and Particle Graph Networks \cite{valencia2019toward} for better tracking the deformation. In a more recent study~\cite{rouhafzay2021transfer}, it has been shown that a deep convolutional neural network for processing vision data can be used with small variations to process tactile data for deformable objects recognition.
	
	Objects made of soft materials, such as human tissues and fruits, have a special force-displacement correlation upon contact. As a result, tactile sensing can be used to estimate the stiffness. In~\cite{yuan2016estimating}, the GelSight~\cite{yuan2017gelsight}, a vision-based high-resolution tactile sensor, measures the 3D geometry of the contact surface, and the normal/shear forces. 
	
	Note that the division of \textit{vision for global deformation} and \textit{tactile sensing for local deformation} is not absolute. The authors of~\cite{guler2014s} present to use vision to estimate the local deformation of objects during grasping, and classify objects accordingly. In~\cite{yuan2018active}, high-resolution tactile sensing is used to estimate the physical properties of clothing materials through squeezing, assuming the robot can learn from the data about indicating global properties of clothing according to a local sampling point. In ~\cite{she2019cable} an example of servoing along a cable based on high-resolution tactile sensing is presented. Although vision is not used, the precise measurement of the local cable shape provides enough information to guide the robot motion on a small scale.  

	\subsection{Challenges and outlook}\label{sec:sensing_chal_n_out}
	Here, the main challenges are: selecting appropriate sensors for the DOM task and using the measurements to obtain meaningful object representations. 
	
	Considering the high number of degrees of freedom (DoF) of the deformable bodies, fusion of different sensing modalities (vision, force and tactile) may be a promising direction to pursue in future research. 
	
	Another research question to be answered is: what yields a good representation of the object configuration? We (acknowledgedly) do not have a complete answer to this; rather, we will elaborate on considerations when designing the representation. 
	
	The representation need to be robust to noise and useful for reconstructing the objects' configuration -- even when data are partially unavailable. In vision, the most common noise is occlusion. How to generate a meaningful representation of these objects under self occlusion is still an open problem in research.
	For rigid objects, one can carefully design the environment to avoid it. For deformable objects that exhibit large global deformation such as clothes, bed sheets etc, self occlusion is inevitable during manipulation. A promising direction to deal with occlusion and noise is using active/interactive perception. With vision data from different perspectives, we might be able to reconstruct the object's configuration accurately even under occlusion and noise.
	
	Apart from the above mentioned challenges, choosing a good representation also involves leveraging two aspects: 
	\begin{enumerate}
		\item the dimensionality of the representation,
		\item the accuracy of the representation.
	\end{enumerate}
	Usually the trade-off depends on the task, relies on human intuition and involves a trial and error process. 
	In end-to-end reinforcement learning settings, sensory data can be mapped directly to robot actions without explicit feature representations \cite{levine2016end}. Human demonstrations can be used for making end-to-end learning more efficient. One example is reported in \cite{matas2018sim}. The authors use an improved version of Deep Deterministic Policy Gradients, trained with $20$ demonstrations, to make robots manipulate cloth. However, since such settings often require a manually designed cost/reward function for learning, human demonstrations in this context can also be used for recovering the reward, with inverse reinforcement learning.
	\section{Modeling}\label{sec:current-model}
	\subsection{Current capability}
	For robots to perform deformation tasks using sensory data, we need a model that captures the relationship between sensor information and robot motion.
	A linear model characterized by Young's modulus can be employed for describing elastic deformation. The two other classes of deformation are: plastic, and elasto-plastic deformations. This classification serves well. Yet, since the model should be used for control, in this section, we prefer to distinguish between local and global models --  a taxonomy which has clearer implications for control. We introduce the corresponding research  and  -- at the end of the section -- we discuss the limitations of these models and present works that address them.
	
	Most local models approximate the perception/action relationship via a Jacobian Matrix (called \textit{Interaction Matrix} in visual servoing). Such a model is linear and can be computed in real-time with a small amount of data. Yet, since it is a local model, it should be continuously updated during task execution. Model updating methods include: Broyden rule \cite{alambeigi2018autonomous}, receding horizon adaption \cite{zhu2021vision}, local gradient descent \cite{navarro2018fourier}, QP-based optimization \cite{lagneau2020automatic}, and Multi-armed Bandit-based methods \cite{McConachie2016mab}. Another advantage of the Jacobian model is that one can design a simple controller by inverting it. However, since this controller is local, it should operate via a series of intermediate target shapes \cite{zhu2021vision}, \cite{lagneau2020automatic}.
	
	On the other hand, global models can be approximated with Finite Element Methods \cite{yoshida2015} and also (deep) neural networks. In contrast to simple linear models, (D)NN-based approaches benefit from stronger representation power, in terms of accuracy and robustness~\cite{valencia2020combining}. Moreover, they can incorporate physics models and reason about object interaction \cite{battaglia2016interaction}. These models can approximate highly nonlinear systems and have a larger validity range, solving (to some extent) the locality issue of the linear models. Nevertheless, these complex nonlinear representations demand large amounts of data (which might not be available in some cases).
	
	Yet, whether we use analytical or learned models, their predictive power will be limited. They are either specialized to some class of tasks or learned from a set of training data. Especially for the learned models, we can never hope to collect enough data to produce an accurate model in the entire state space (which is high dimensional). Thus \cite{McConachie2020} and \cite{Power2021} have developed methods to reason about the validity of a (learned) model for a given state and action, and have used these methods to reason about model uncertainty in planning and control. However, when the model is not precise, a re-planning/recovery might be desirable. The authors of \cite{mitrano2021learning} introduces two neural networks for learning and re-planning the motion when the model is unreliable.   
	\subsection{Challenges and outlook}\label{sec:model-outlook}
	The complexity of modeling is manifested in the lack of simulators. While most existing robotic simulators are capable of producing rigid body kinematics and dynamic behaviours, only a fraction of them can handle deformation. One recent work, Softgym \cite{lin2020softgym} was proposed for bench-marking DOM based on Nvidia Flex. In the soft robotics community, SOFA \cite{coevoet2017software} and Chainqueen \cite{hu2019chainqueen} are example simulators. In Sect. \ref{sec:hardware_outlook}, we considered the interaction between soft robots and deformable objects. Thus, a unified simulator that is able to handle soft robots and objects, and model their interaction might be desirable.
	
	When choosing a model for control, one challenge of data-driven deformation modeling is to balance region of validity with number of data required for training. One possible direction is to combine a simple model with a complex nonlinear model to form a hierarchical model. An example of such structures is exploited in \cite{li2020learning} for robust in-hand manipulation.
	For DOM tasks, we can have a linear model at lower level, and a (D)NNs learning the full model at higher level. 
	The lower level model can be learned in few iterations to enable instant interaction between robot and object. The higher level (D)NN can collect data and improve the model to enhance global convergence.    
	
	\section{Planning}
	\subsection{Current capability}
	Planning aims at finding a sequence of valid (robot/object) configurations and contributes to solving the problem of limited validity of local models, as discussed in Sect. \ref{sec:current-model}. 
	
	Planners can operate in the objects' configuration space, and sometimes rely heavily on physic-based simulation. While the obtained plan can be visually plausible, it may be unrealizable for a specific object. Recently, McConachie et. al. presented a framework which combines global planning without physics simulation, with local control~\cite{mcconachie2020manipulating}. For an elastic object, considering its energy is another way to do planning; in this direction, Ramirez-Alpizar et al. \cite{ramirez2014} proposed a dual-arm manipulation planner optimizing the elastic energy, for elastic ring-shaped objects manipulation. 
	For DOM tasks involving multiple robots, planning is important for coordination. Alonso-Mora et al. employed a distributed receding horizon planner for transporting tasks that require multiple robots \cite{alonso2015local}. More recently, \cite{9410363} learns a latent representation for semantic soft object manipulation that enables (quasi) shape planning with deformable objects.
	
	With Learning from Demonstration (LfD), the robot can be trained to manipulate deformable objects by an expert (usually a human). LfD encodes the robot trajectory and interaction force from human demonstrations \cite{lee2015learning_1}, thus avoiding explicitly planning the motion. More recently, Wu et al. have proposed a reinforcement learning scheme for DOM, which does not require initial demonstrations \cite{wu2019learning}. 
	\subsection{Challenges and outlook}
	A rigid object configuration can be described in space with 6 DoF, whereas a deformable object configuration has a much higher number of DoF. To address this from the sensing algorithm side, one can find a compact representation from sensory data, as discussed in Sect. \ref{sec:sensing_chal_n_out}. An alternative, which receives much less attention, is the use of environmental contacts to constrain some DoF of deformable objects. Examples include the use of contact points in cable harness or that of flat surfaces when folding clothes. We argue that instead of planning to avoid contacts as most planners do, for deformable objects, we need to plan to make contact, since this constrains the configuration, and therefore simplifies the task.
	
	Planning to grasp the correct point is often crucial in DOM tasks. For instance, grasping at convex vertices of the clothes guarantees stability and facilitates the task~\cite{van2010gravity}. Re-grasp planning is highly relevant when considering tasks which require multiple robotic arms. Additional challenges come from perception, since as soon as the robot releases one or more grasp(s), the object is likely to change its configuration. We rely on sensing to track configuration changes and then plan accordingly. 
	
	Another important future work in planning is reasoning about a deformable object at a semantic level. What does it mean for a cloth to be \textit{folded}? What does it mean for an object to be \textit{wrapped} in paper? We cannot manually specify all the configurations of the deformable object to use as goals in these kinds of tasks. Instead, we need a way to learn the meaning of semantic concepts, such as \textit{folded} or \textit{wrapped}, so that we can determine if a given configuration of the object is a valid goal.
	
	\section{Control}\label{sec:control}
	
	\subsection{Current capability}
	
	Control aims at designing inputs for the robot to realize the planned motion. The type of controllers is decided usually by the task. For instance, the authors employed a data-driven model predictive control \cite{9035011} for cutting considering its predictive nature and the lower demand for manual tuning. For safe interaction in minimally invasive surgery, the authors of \cite{su2019improved} used a fuzzy compensator with impedance control. For controlling large deformation, Aranda et al., proposed a Shape-from-Template algorithm concerning its low dimensional representation (using the template) and robustness against occlusion \cite{aranda2020monocular}.
	
	A number of works focus on shape control.
	While global models directly map sensor data to robot motion, local models must be inverted to design the robot motion controller (see Sec.~\ref{sec:current-model}). 
	Several applications of the control scheme for robotic manipulation of deformable objects can be found in 3C manufacturing \cite{li2018pcb,li2018usb}, where vision-based controllers were proposed to drive the robot to automatically grasp/contact the deformable object, then carry out the task of active deformation or separation/sorting.
	Other works consider the concept of diminishing rigidity to do deformation control \cite{nadon2020grasp}, \cite{ruan2018accounting}.
	
	\subsection{Challenges and outlook}
	
	Feedback control has been commonly used in most DOM works, by referring to the state of the object, to achieve the task. Note that such state is retrieved from the output of its deformation model and measured with sensors, and that output and state do not necessarily have the same representation and dimension.
	Furthermore, we can distinguish between model-based and model-free control. Due to the complexity of modeling the deformation, when using the model to derive control policies, the controller has to take into account that the model will be inaccurate or even wrong. 
	
	
	Model-free approaches do not require information about the deformation parameters or the structure of the deformation model. 
	Examples include LfD or (Deep) reinforcement learning, where the challenges are: efficient use of data, and policy generalization. To address these issues, we can combine the offline and online learning methods. In the offline phase, the supervised network can be trained to estimate the model, by collecting pairs of a series of predefined inputs (e.g., the velocity of the robot end-effector) and the deformation of the object. The estimated model in the offline phase can be further updated online during the control task with adaption techniques (e.g., the adaptive NNs), to compensate the errors due to insufficient training in the offline phase or the changes of the deformation model. Hence, both complement each other.
	
	When multiple features on the deformable object are controlled in parallel, the system becomes under-actuated, with less control inputs than error outputs. Then, the robot controller should be able to deal with the conflicts between multiple features or decouple the control of multiple features in a sequential manner, to guarantee controllability.
	
	In addition, due to the deformation during control, the contact between robot end-effector and deformable object may not always be maintained. Most existing systems require a certain level of human assistance to initiate the contact or to re-establish it, if it is lost during the task. To improve autonomy, the robot controller should automatically grasp or touch the object first, whenever physical contact is lost, laying the foundation of the subsequent manipulation task. Such a capability would allow the robot to effectively deal with the unforeseen changes due to deformation.

	\section{Practical applications}\label{sec:application}
	In previous sections, we centered our discussions from a scientific point of view, here, we instead discuss challenges in various applications where DOM can be translated to solutions. 
	
\begin{figure}
		\centering
		\includegraphics[width=0.48\textwidth]{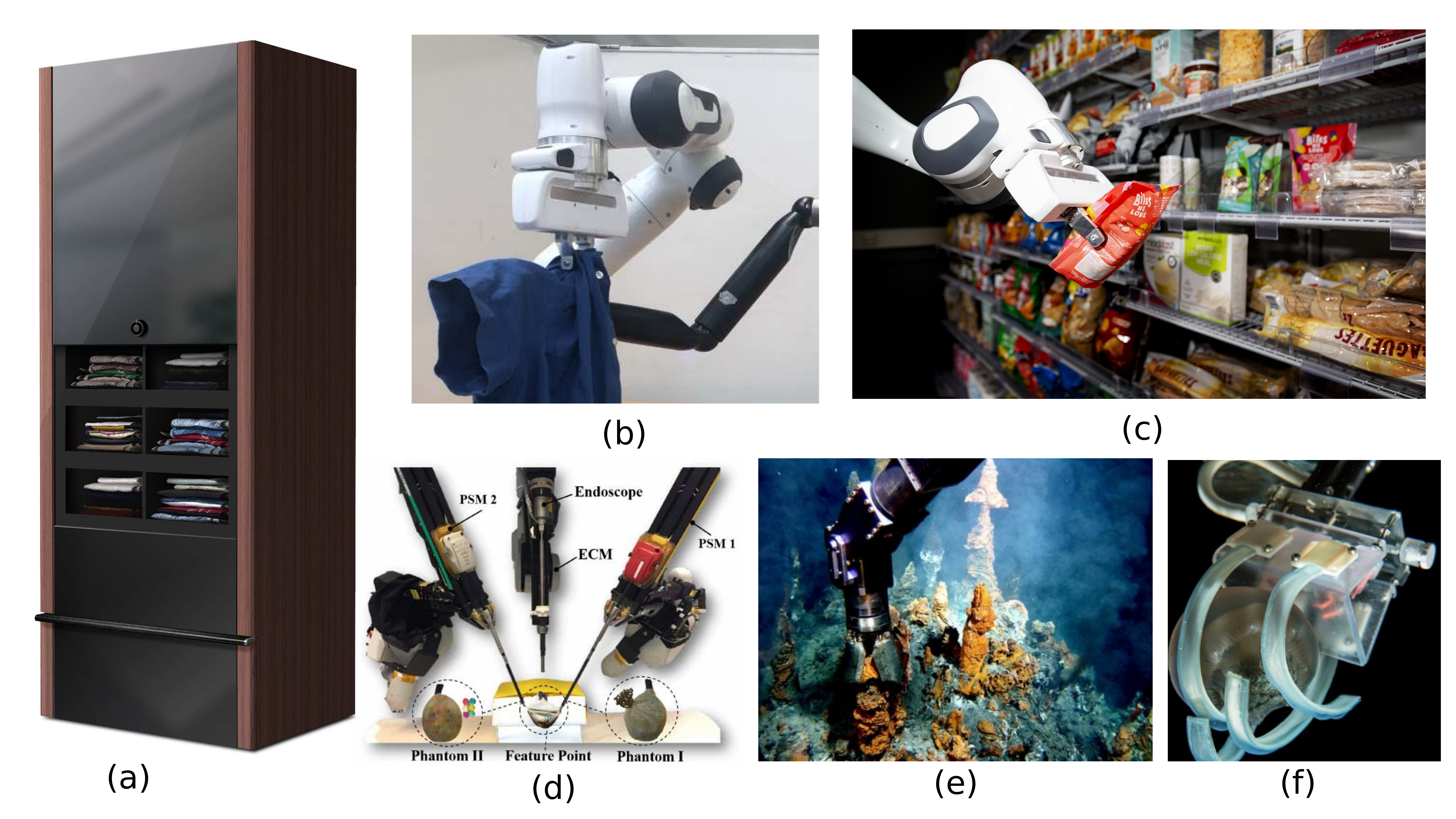}
		\caption{Various applications of DOM -- (a): laundry-folding robot from Seven Dreamers Laboratories Inc. \cite{laundry}, (b): A mock-up for robotics dressing assistance, (c): a robot picking a flexible bag of goods on the shelf, courtesy AIRLab Delft \cite{AIRLab}, (d): autonomous surgical manipulation by the dVRK system \cite{alambeigi2018robust}, (e): ROV Victor 6000 sampling black smokers (IFREMER/GENAVIR) courtesy D. Desbruyères, (f):  Ultra soft underwater gripper for jellyfish \cite{Sinatra2019-sr}}.
		\label{fig:applications}
	\end{figure}

	\textbf{Automatic laundry:} A typical domestic application of DOM is laundry folding. A Tokyo-based company unveiled its prototype laundry-folding robot in 2015 (Fig. \ref{fig:applications}a). However, the company was announced bankrupt in 2019 due to lack of funding for development and difficulties in improving the robot to reach a satisfactory level \cite{nagata_2019}. Although cloth folding has been tackled in a few previous research \cite{verleysen2020video,miller2012geometric,tsurumine2019deep,borras2020grasping}, it remains largely a laboratory product (limited to structured environments, certain types of the clothes, etc). Commercializing the technology seems requiring a substantial efforts.
	
	\textbf{Assistive dressing:} Robotic dressing assistance has the potential to become an important technology due to the pressing needs for ageing society support. 
	Research can roughly be categorized into simulation-based learning \cite{clegg2018learning,clegg2020learning} and imitation learning \cite{joshi2019framework} approaches. Examples are dressing support for shoes \cite{canal2019adapting}, shirts \cite{Li-RSS-21,tamei2011reinforcement,zhang2019probabilistic} and pants. 
	However, several technical and societal challenges have to be addressed before robot-assisted dressing will become a broadly used DOM technology: physical safety for the human, modeling and prediction of the human-robot interaction, robustness for large variations of geometric and dynamic properties of textiles, low-cost high-reliable robot hardware, human acceptance of such technologies.
	
	\textbf{Surgical robotics:} Soft tissue manipulation is mainly performed with tele-operation solely using visual feedback. Autonomous manipulation, however, still has a long way to go and demands developing various DOM hardware and software (Fig. \ref{fig:applications}d). The biggest concern for an autonomous solution is the safety of operation. A soft robot with intrinsic compliance will probably enhance the safety.
	
	
	\textbf{Food production \& Retail:} Handling deformable objects is a major challenge in the whole chain from production to sales. In an agricultural setting, automated harvesting of fruits and vegetables requires interactions with deformable objects that are at the same time easy to damage, which immediately decreases their value and shelf live. Frequently, these  products also undergo an intermediate processing step (e.g., filleting and packaging meat). More generally, deformable products (e.g., everything packaged in flexible bags, (Fig. \ref{fig:applications}c)) need to be handled in warehouses, in order picking, and in restocking. Solutions for specific applications and products have been developed, but more complex objects and operations still are frequently handled by human workers. 
	
	\textbf{Marine robotics:} Underwater grasping has been led by oil and gas industry for decades, resulting in heavy machines with strong grippers for inspection and maintenance tasks (Fig. \ref{fig:applications}e). Gradually the demands turned to more detailed tasks in marine biology, sedimentology and archaeology (Fig. \ref{fig:applications}f). Another DOM application can be found in tethered robot umbilical modeling and control. Negative buoyancy cable can be modeled in real time as a simple catenary shape and tracked to control a tethered ROV~\cite{LARANJEIRA2020107018}. 

	\section{Summary and key messages}\label{sec:summary}
	
\begin{figure}
		\centering
		\includegraphics[width = 0.49\textwidth]{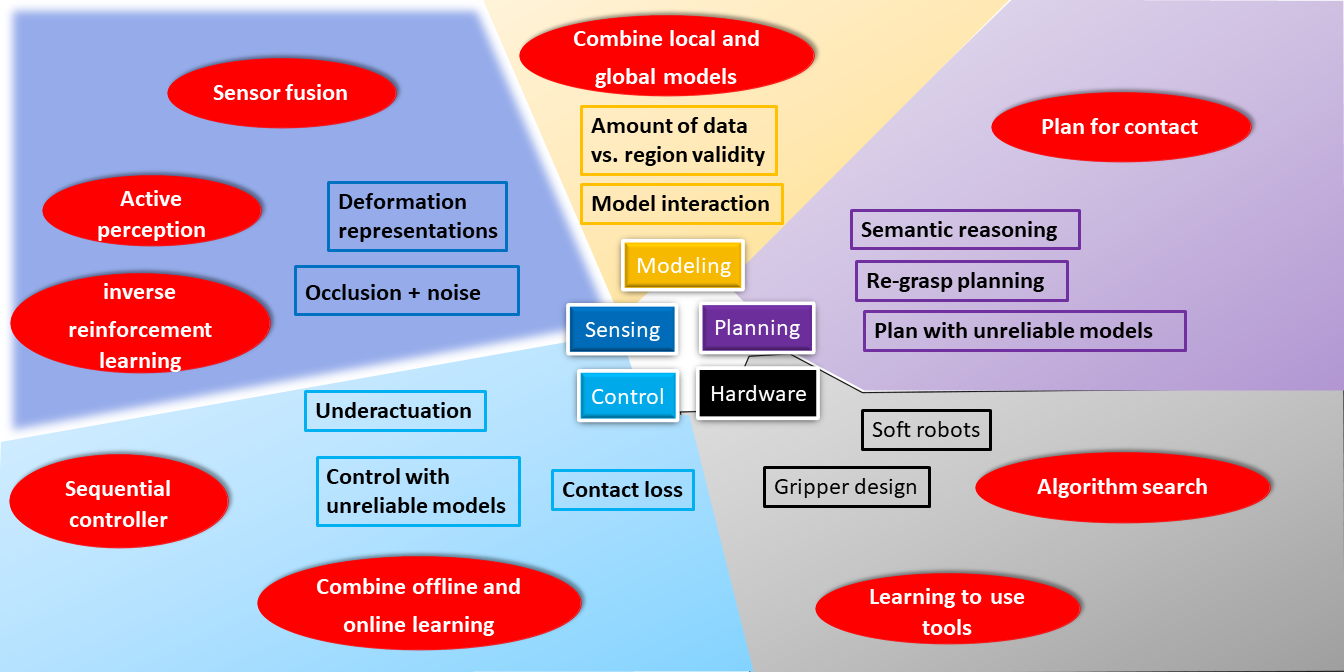}
		\caption{A summary of open research problems and ideas/methods to pursuit discussed in this paper. Research problems in each subarea are written with bold black texts, whereas ideas/methods to resolve them are marked with bold white texts in red eclipses.}
		\label{fig:DOM_challenges}
	\end{figure}

	The revolution of robots from automating repetitive tasks to humanizing robot behaviours is taking place with better hardware,  robust sensing capabilities, accurate modeling, increasingly versatile planning and advanced control. Manipulation of deformable objects breaks fundamental assumptions in robotics such as rigidity, known dynamics models and low dimensional state space. It therefore requires breakthroughs in all the areas mentioned above, and serves as a great test-bench for novel ideas in both robotic hardware and software. A summary of challenges and ideas discussed are presented in Fig. \ref{fig:DOM_challenges}.
	
	In terms of hardware, recently, the community has been shifting more and more from rigid to soft robots. Robotic manipulation is also gradually shifting from rigid to deformable objects. One open question is if some of the algorithms in one field are transferable to the other? We believe the interaction between a soft robot and a deformable object will bring more challenges to the robotic community.
	
	Sensing plays a vital part in robotics manipulation of deformable objects. Depending on the nature and complexity of the task, one or multiple fused sensing modes may be needed. Machine learning will facilitate the development of robust algorithms to process data from different sensors, to generate meaningful representations of deformation.
	
	\emph{All models are wrong, some are useful}. We do not believe there exists the \say{best} model for deformation. While more and more models tend to be data-driven, we would like to draw the readers' attention to the importance of physical models for studying interactions.
	
	For planning, current research lacks a high level semantic reasoning of the DOM task. Furthermore, while often the purpose of planning is to avoid contact and collision, we argue that for DOM, it can be very useful to plan for contact.  
	
	Under-actuation is a key challenge of DOM, due to the deformable bodies' high DoF. Another practical issue introduced with deformation is contact loss during manipulation; future controllers should be able to detect contact loss and to react accordingly.

	
	\bibliography{reference.bib}
	\bibliographystyle{IEEEtran}
\end{document}